# Sexing Caucasian 2D footprints using convolutional neural networks

Marcin Budka[1¶], Matthew R. Bennet[2¶*], Sally Reynolds[3] Shelby Barefoot[2] Sarah Reel[4&], Selina Reidy[5&], Jeremy Walker[6&]

[1]Department of Computing and Informatics, Bournemouth University, Poole, BH12 5BB, United Kingdom.

[2]Department of Environmental and Life Sciences, Bournemouth University, Poole, BH12 5BB, United Kingdom.

[3]Department of Archaeology and Anthropology, Bournemouth University, Poole, BH12 5BB, United Kingdom.

[4]Division of Podiatry and Clinical Sciences, University of Huddersfield, Queensgate, Huddersfield, HD1 3DH, United Kingdom.

[5]Identification Bureau, Yorkshire and the Humber Regional Scientific Support Services, Sir Alec Jeffreys Building, Peel Avenue, Calder Park, Wakefield, WF2 7UA, United Kingdom.

[6]Sheffield Teaching Hospital, NHS Trust, Podiatry Services, Woodhouse Clinic, 3 Skelton Lane, Sheffield, S13 7LY, United Kingdom.

* Corresponding author

Email: mbennett@bournemouth.ac.uk

[¶] These authors contributed equally to this work.

[&] These authors contributed equally to this work.

# Abstract

Footprints are left, or obtained, in a variety of scenarios from crime scenes to anthropological investigations. Determining the sex of a footprint can be useful in screening such impressions and attempts have been made to do so using single or multi landmark distances, shape analyses and via the density of friction ridges. Here we explore the relative importance of different components in sexing two-dimensional foot impressions namely, size, shape and texture. We use a machine learning approach and compare this to more traditional methods of discrimination. Two datasets are used, a pilot data set collected from students at Bournemouth University (N=196) and a larger data set collected by podiatrists at Sheffield NHS Teaching Hospital (N=2677). Our convolutional neural network can sex a footprint with accuracy of around 90% on a test set of N=267 footprint images using all image components, which is better than an expert can achieve. However, the quality of the impressions impacts on this success rate, but the results are promising and in time it may be possible to create an automated screening algorithm in which practitioners of whatever sort (medical or forensic) can obtain a first order sexing of a two-dimensional footprint.

# Introduction

Within the forensic, podiatry and anthropological literature there have been several attempts to determine the sex of bare two-dimensional (2D) human footprints [1]. The aims and justification for these studies are varied and range from basic medical and anthropological description, via victim profiling as an aid in disaster identification, to use in criminal forensic casework [2]. In the latter context, 2D footprints are often found at crime scenes where bodily fluids have been tracked around a scene via bare feet or those encased in socks. Most forensic practitioners engaged in the preparation of judicial evidence, and/or its review, would currently refrain from determining footprint sex. This is due primarily to a lack of reliable tools for sex determination and the risk that false intelligence might be provided. If a tool could be shown to be reliable and universal, then this might change. The application of machine learning may provide such a tool and this forms the aim and focus of this paper [3]. This paper therefore focuses on an application of machine learning via the use of convoluted neural networks (CNN) rather than advancing the development and fine tuning of such algorithms and it is the application which is both novel and of interest here.

Our ability to sex a footprint is a function of the degree of sexual dimorphism in our species. Most species in the animal kingdom have larger females than males with the exception of mammals and birds [4]. Variation in dimorphism with body size follows the so-called Rensch's rule [5] in which taxa with larger males show greater dimorphism (hyperallometry). Human sexual dimorphism is best illustrated by stature within males who are typically 7% taller, and since stature and foot length correlate well, as indicated by the numerous empirical relationships [1], it should also be manifest in footprint length [6]. Dimorphism emerges primarily during postnatal growth, with male neonates only 1% longer than females at birth [7]. During childhood, dimorphism remains relatively minor until the onset of puberty when it becomes established. Sexing a footprint is therefore restricted to adult cases. In a forensic context this is complicated by variation between and within populations. Rodriguez et al. (2005) suggests that dimorphism is a function of phylogeny and selection pressures such as marriage systems [8][9][10][11], social stratification [10], sexual division of labour [11] and potentially nutritional standards [9,12]. Variation in stature with climate, as predicted by Bergmann's rule [13] has previously received some empirical support (e.g., [14,15]) and could potentially be a source of variance. The net result of this is that empirical relationships are likely to be subject to inter-population variance and when coupled with the modest (i.e., 7%) level of sexual dimorphism in *Homo sapiens* will limit the degree of discrimination that is possible.

The sexing of footprints has primarily proceeded by comparison of linear dimensions (e.g., [16–21]) and to a lesser extent via shape metrics (e.g.,[22,23]). There is a known sex difference in the density of friction ridges on human fingers [24] which has been translated to feet in recent studies (e.g., [25–27]). Differences have also been observed in skeletal elements namely the tarsal and calcaneus bones (e.g., [3,28–34]). Here we develop a machine learning approach which looks at the entirety of a footprint in an attempt to determine its sex. We then partition discrimination between surface texture, inclusive of frictions ridges, and overall morphology.

# Methods and materials

## Data sources

A pilot study was first completed using data collected from student volunteers at Bournemouth University in the autumn of 2018. Volunteers left a standing footprint using an inkless pad system and were asked to confirm the sex they were registered as at birth. A total of 101 males and 132 female right footprints were obtained from adults between the ages of 18 and 25 years old and predominantly, although not exclusively white Caucasian. Difference in footprint size between traces left during walking (so-called dynamic traces) and those left in static tests are also well documented [35] and moreover one might expect a difference in the quality of friction ridges left between dynamic and static traces. In this initial study static traces were used for ease of execution, but clearly this something that should be factored into future research as discussed later. Footprint impressions were scanned and saved anonymously for analysis in accordance with ethical approval obtained from Bournemouth University (ID: 22317) and informed consent of the volunteers. A series of digital landmarks [1] identifying the main dimensional properties of the foot were placed digitally by an operator on each footprint image using DigTrace (www.digtrace.ac.uk) (Fig. 1a). Landmarks were exported as coordinates, and these were used to compute linear dimensions. A script was written to place 10 mm$^2$ sampling squares between selected landmarks and used to crop the image into a series of individual image fragments. The black and white pixel proportions were calculated for each of the squares and the ridge density was counted manually and a 10% sample verified or accuracy. The geometric morphometric analysis and discriminant analyses conducted on the landmark data were undertaken in the statistical software package PAST [36].

The author received additional data in the form of 2677 footprint images provided by Jeremy Walker of Sheffield NHS Trust, hereby referred to as the Walker-Data. This data was received in a fully anonymised state and subject's provided informed ethical consent for their anonymised data to be used in research. The original ethical approval was provided by the Sheffield NHS Trust. The data consisted of 2677 binary .tif images with 2240 x 3200 resolution at 200 ppi, of either a right or left print for 1483 females and 1194 males, aged between 16 and 81 with over 97.5% being Caucasians. Landmarks were placed digitally by an operator using the freeware DigTrace (www.DigTrace.co.uk) on a training set of 200 randomly sampled from the 2677 population. This training set was used to train an automated landmark placement algorithm and applied to the remaining population. Sampling squares were obtained in the same way as for the pilot project however ridge density was not manually counted. A total of 20% of the images were selected at random and isolated for validation purposes. The images were cropped to a bounding box to eliminate background noise which has the potential to leak information. In order to reduce the computational requirements, the images were resized to 512x640. This aspect ratio is different from the original images as it better reflects the shape of the bounding boxes in the dataset; the aspect ratio of the footprints themselves was not altered. The sampling algorithm impacts on the information encoded within each image. The common default of BILINEAR interpolation results in loss of ridge detail but makes the texture appear softer and therefore more representative of a photographic image. From the other standard image down-sampling techniques, NEAREST neighbour preserves the binary, black and white, nature of the original images while HAMMING produces a sharper image than BILINEAR and does not have the dislocations at a local level that BOX does (Fig. 2). As a result of sensitivity testing and with the aim of maximising the information available, NEAREST neighbour, BILINEAR interpolation and HAMMING were the three channels selected of the same input image (Fig. 3a). An alternative texture-free version of the image as shown in Figure 3b, was created using erosion followed by dilation (fundamental morphological image processing operations) for 5 iterations each, with a 3 x 3 kernel. All the footprint images and associated metadata are available for the purposes of replication from Bournemouth University's Data Repository (https://doi.org/10.18746/bmth.data.00000157).

## Machine learning algorithms

Over the past eight years there has been a rapid increase in applications of Convolution Neural Network (CNN) to a range of applications ranging from medical imaging [e.g., 37,38] to the analysis of forensic footwear impressions [39] and even the detection of building defects [40] to name just a few.

In order to build a predictive model for sex estimation (and age as a by-product), we experimented with several CNN architectures, namely ResNet34, ResNet50 and ResNet101 [41], pre-trained on the ImageNet2012 dataset. The input images have been resized to 512x640 pixels as discussed in the previous section. We have built a separate model for the following three tasks: (1) sex estimation; (2)

age estimation; and (3) combined sex and age estimation. For task (1) the network had a single output neuron producing an estimated probability of the print being female via the sigmoid function, and we have used the binary-cross entropy loss:

$$\mathcal{L}_c = -(y_c \log \hat{y}_c + (1 - y_c) \log(1 - \hat{y}_c)) \quad (1)$$

where $y_c \in \{0, 1\}$ denotes the label and $\hat{y}_c \in \langle 0,1 \rangle$ is the network output.

For task (2) the network also had a single output neuron, this time producing an unconstrained age estimate, and was trained using the L1 loss:

$$\mathcal{L}_r = |y_r - \hat{y}_r| \quad (2)$$

where $y_r$ is the actual age and $\hat{y}_r$ is the network output or prediction.

For task (3) there were two outputs and we have used a weighted sum of the above two loss functions:

$$\mathcal{L} = \mathcal{L}_r + \lambda \cdot \mathcal{L}_c \quad (3)$$

where $\lambda$ has been empirically set to 20.

For each task, we have replaced the final fully-connected layer from the pre-trained network with a custom head consisting of two linear layers with ReLU [42] non-linearity in-between, and used dropout [43] and batch normalisation [44]. We first trained the head only keeping the rest of the network fixed for 10 epochs using the Adam optimiser [45], and subsequently fine-tuned the whole network for another 10 epochs. One of the initial challenges of this model was excluding non-pertinent information. Of note, in an initial run of the model using the Bournemouth-Data it found a way of effectively cheating. The slip with the print code and sex was left attached initially with the whole image being used in its raw state. Figure 4 illustrates how the model quickly identified the significance of the tick placement denoting the sex of the participant.

In terms of evaluating the model there are a range of approaches. Cross-validation has been the gold standard in machine learning for several years [44,46]. This involves splitting the data into K approximately equal parts, training the model on K-1 parts and testing on the remaining part. This is then repeated K times always leaving a different part for testing. However, since the resurgence of deep neural networks [47] this has changed in favour of a hold-out method in which a single random split, usually 80-20, is used [41]. This is the approach we have used in the current work. One of the issues cross-validation aimed to address was the problem of small datasets, where a hold-out set would be unlikely to be representative of the data distribution as a whole. With more than 2600 examples in the Walker-Data, with all being bare footprints collected in exactly the same way, this risk is reduced. In addition, the computational cost of training K models rather than 1 for each experimental setup is prohibitive for most modern CNN architectures. Finally, the vast majority of seminal neural network papers published in the last 7-8 years (e.g., [41,43,44]) do not use cross-validation for the similar reasons to those outlined above. We do acknowledge however that if our work were to be used in jurisprudence, more rigorous evaluation would be required.

# Results

## Conventional landmark analysis

Using both the Bournemouth and Walker data and the placed landmarks (Fig. 1) conventional geometric morphometric analyses were undertaken [1]. Figure 5 shows the distribution of landmarks in the dataset following a Generalised Procrustes Analysis (GPA) and the thin-plate spline for the male compared to female landmark means for the Bournemouth-Data. Less than 40% of the landmark variation is accounted for by the geometry of the longitudinal media arch, with arch being slightly less well-defined in the female footprints. Linear Discriminant Analysis (LDA) gives an initial sex discrimination of 76.92%, which when jack-knifed falls to 65.81% (Table 1). For the Walker-Data the results are similar 70.95% falling to 69.34% when jack-knifed (Table 2). Using the 153 possible inter-landmark distances without any GPA and therefore incorporating aspects of both shape and size the LDA gives an initial sex discrimination of 98.71%; reduced to 67.38% when jack-knifed (Table 1).

## Friction ridges

Using the Bournemouth-Data friction ridges were used to examine their ability to sex the 2D footprints. Sample squares were initially cropped from the images automatically between equidistant selected landmarks (Fig 1a). The percentage of black versus white was computed automatically for each square using a high contrast version. In theory the greater the percentage of black the greater the ridge density should be. This resulted in a discrimination of only 59% falling to 53% when jack-knifed. Discrimination values are reported in Table 1. The results are relatively poor which could be a function of the callus build in some areas of the foot, or the poor quality of some of the footprint images in areas of maximum plantar pressure. A second set of sample squares were extracted to avoid 'high contact' areas of the foot (Fig. 1b) and the analysis was repeated. The results obtained from the automated black-white percentage were better but still relatively poor. Ridge counting for each sampled image square followed the procedure of Krishan et al. [25] and [26] was then used as a further test. Using the data for 168 participants, 29 footprints were excluded due to being faint, an average total ridge count for females were obtained of 52 versus 47 for males across the three squares. Furthermore, the average ridge count of each individual square was higher for females than males. A discriminant analysis gave a 67% success rate (Table 1).

## Machine learning

Table presents results for the five possible combinations of data input, namely: (1) Shape + Texture + Size; (2) Shape + Texture; (3) Shape + Size; (4) Shape only; and (5) Texture only. Inclusion of size is only possible in conjunction with shape and the texture only version is based on the sampling of squares from the footprint as shown in Figure 1. The aim here was to explore how important different elements of the impression where in making the classification. Each of these five inputs has been used for three different tasks: sex estimation, age estimation model, and combined sex and age estimation. This analysis was run on the Walker-Data being the larger of the data sets and using ResNet101 as it consequently outperformed the other variants tested. For all the experiments a random, 80/10/10 training/validation/test split of the dataset was used, and the metrics reported in the table apply to the test set.

The maximum discrimination that can be obtained is just under 90% with all elements (size, shape, and texture) used in the analysis; that is the entire image. Interestingly 83% is achieved for using texture alone but excluding it does not materially reduce the power of size or shape. Surprisingly, the best results are obtained when the model learns to estimate sex and age collectively. Although it might seem counterintuitive at first, it is possible that the richer feedback signal coming from two different sources instead of one helps the model converge on a better result. Therefore, improvement in one of the subtasks does not necessarily compromise the performance in the other. The learnt intermediate representations of the patterns are probably also more semantically meaningful as the model is less likely to learn any idiosyncrasies associated with the individual subtasks. While we remain uncertain what the model is 'seeing' in the impression to estimate age, and therefore would treat this data with considerable caution, setting the model to estimate this as well helps improve the overall result.

The textured images within the Walker-Data footprints, scaled to preserve their relative size (Scenarios 1, 6 and 11, Table 2), were sexed correctly almost 90% of the time. Figure 6 shows the heat maps associated with the most successful classifications (i.e., the ones with the highest confidence). The model was also able to predict the age of the footprint-maker to a mean absolute error of 7.51 years. Repeating the analysis with the texture-free version of the footprints (Scenarios 2, 7 and 12, Table 2) reduces the classification accuracy only slightly to 89.13% and increases the age error to 8.79 years. This suggests that the majority of the discrimination is achieved via gross variation in size and shape of the foot images, and to a lesser extent by the texture associated with such things as ridge density. The proportion to which discrimination is partitioned between texture and gross morphology will vary with the resolution of the input images. For example, repeating the exercise using the Bournemouth-Data which has different image resolution gives a textured discrimination of 82.6% and a non-textured discrimination of 65%. The ridge detail is more prominent in the Bournemouth-Data and consequently appears to be doing a greater proportion of the work.

Comparison of the results for Scenarios 7 and 8 (i.e., size vs texture) reinforces this observation even further, replacing texture with size results in over 4% accuracy boost, although at the expense of the quality of age estimations. However, the Shape vs Texture comparison (Scenarios 9 and 10, Table 2) hints towards the latter having more discriminative power (83.89% vs 80.15%), once again at the expense of age estimation. The CNN outperforms conventional landmark-based analysis that involves shape only (Table 2, Scenario 16) but is equivalent to an inter-landmark analysis (shape + size; Table 2, Scenario 17). Manual landmark placement would make the CNN preferably due to the time saving, however it is worth noting that landmark placement was automated for the Walker-Data so in either case machine learning gives slightly better performance overall.

The heat maps provide an opportunity to examine which areas are most critical in determining the classification of a specific individual. Highlighted areas correspond to areas where a small change would impact on the classification (Fig. 6). Essentially this allows us to determine which morphological areas are doing the most 'discriminative-work'. Two areas seem to be key one between the first and second toes and a second around the heel (Fig. 6). The later makes sense since the width of the calcaneus/talus is an established measure of sexing the bones of the foot (e.g., [31,34,48,49]). The former is less easily interpreted, although features in the toe region (e.g., humps, phalange marks) have been identified as highly individual [21]. The shape in this region may also be influenced by gender based shoe choices[50].

## Discussion and conclusion

This research demonstrates that accuracies of over 85% are possible when sexing footprints using a 'whole-foot' machine learning approach in which size, shape and texture are included in the model. This has been achieved with a relatively modest data library for training purposes, and improvement is no doubt possible with the availability of increased data volume. In the current model, the relative work done in achieving discrimination varies between the various image-components and is dependent on the image quality of the impressions used. The more texture (i.e., ridge detail) that the footprint images contain, the more that this aids sex discrimination. In a forensic application this is often obscured by socks and friction ridges may be obscured/clogged by blood or other fluid. Extraneous noise around the image may also be an issue. The current model should be considered as baseline only that can be expanded, and made more robust, by the addition of more training data across a range of potential applications. It shows potential however given more training data to improve its applicability.

Surprisingly, the model demonstrated some ability to age footprints as well, there is uncertainty however about what the model is actually 'seeing' unlike sexing, which seems to focus on the shape of the heel. Therefore, at this stage these results should be treated with considerable caution. However, setting the model to estimate age as well as sex helps improve the overall result. Feet stop growing following maturity therefore past this point, natural growth of the foot alone cannot be responsible for the machine's ability to age a footprint. However, changes to the feet may still occur in maturity due to systemic disease, musculoskeletal pathologies, and trauma. Furthermore, the skin can atrophy and become anhidrotic with age and pathology whilst plantar subcutaneous fat may also reduce. Whether the model is 'seeing' these changes remains uncertain. Irrespective of the validity of the age estimates it is clear that improved sex discrimination is achieved by setting the machine learning model multiple tasks of which age is an obvious one.

In terms of applicability of this model and approach are varied provided that the model can be made robust enough to take a range of inputs. Forensic podiatry is an obvious future application. Currently forensic podiatry casework focuses on identifying similarity and dissimilarity between footprints left at a scene (questioned footprints) and sample footprints made by a suspect (reference footprints). This requires both questioned and reference footprints to be available for evidential comparison. In some circumstances, only the questioned footprint may be available, and its analysis may assist the investigation if accurate intelligence regarding the sex or age of potential suspects could be provided. In other scenarios however a first order assessment of sex may be appropriate for example in victim identification or rapid profiling to aid investigative direction.

As stated above the results look promising but a lot more training data of various types is needed to take the next step and produce something which can be deployed operationally. There are lots of issues to consider, of which the following are just a few. Most footprint examiners, at least in the UK, see partial marks so an operational model in the future will need to be able to give results on these. It should be possible in the short term to create 'partial prints' by cropping the existing training data and this is an obvious early win. However, the model has been trained with static prints, whereas most traces at crime scenes are dynamic prints (i.e., walking or running). Therefore 'real' forensic traces need to be obtained to train the model further. In addition, twisting and other forms of natural distortion all need to be considered. Recovered prints can also be presented as negatives and a future model will need to be able to invert these types of images for analysis. The current model works with either a right or a left foot and has yet to be applied to a trackway of prints. In theory sex determinations made on multiple tracks of the same person should increase the reliability of the conclusion. The final limitation of the current model is that both the data sets used are dominated by one racial group and an ethnically diverse training population would help. To amass the training data that is needed we need a community-based approach to this in which data is sourced from across the world in a variety of forms for training purposes. The sheer number of publications over the last decade which use two-dimensional print images is huge and illustrates the potential data that could be harnessed to develop a future app for use by everyone. One final point worth stressing is that should the approach gather momentum with the addition of training data and before it is used in jurisprudence it would also be essential to explore different routes for validation, including perhaps the computationally intensive cross validation approached considered to be some as the gold standard.

# Tables + Captions

| Bournemouth Data | Protocol | % Discrimination | Jack-Knifed |
|---|---|---|---|
| N=233 | Landmark coordinates (Fig. 1a) | 76.92% | 65.81% |
| N=233 | Inter-landmark distances (Fig. 1a) | 98.71% | 67.38% |
| N=233 | Binary % black, seven squares (Fig. 1a) | 59.13% | 53.48% |
| N=233 | Binary % black, three squares (Fig. 1b) | 50.64% | 53.48% |
| N=99 | Manual ridge counting three squares (Fig. 1b) | 67% | 43.35% |
| N=233 | Machine Learning - Shape+Texture+Size | 82.6% | - |

Table 1: Sex discrimination on the Bournemouth data using various methods.

| Scenario | Task | Shape | Texture | Size | Accuracy (Sex) | MAE (Age) |
|---|---|---|---|---|---|---|
| 1 | Sex estimation | Yes | Yes | Yes | 88.76% | - |
| 2 | Sex estimation | Yes | No | Yes | 88.76% | - |
| 3 | Sex estimation | Yes | Yes | No | 79.40% | - |
| 4 | Sex estimation | Yes | No | No | 78.27% | - |
| 5 | Sex estimation | No | Yes | No | 80.15% | - |
| 6 | Sex & age estimation | Yes | Yes | Yes | 89.89% | 7.51 |
| 7 | Sex & age estimation | Yes | No | Yes | 89.13% | 8.79 |
| 8 | Sex & age estimation | Yes | Yes | No | 85.01% | 8.26 |
| 9 | Sex & age estimation | Yes | No | No | 80.15% | 8.98 |
| 10 | Sex & age estimation | No | Yes | No | 83.89% | 9.20 |
| 11 | Age estimation | Yes | Yes | Yes | - | 8.47 |
| 12 | Age estimation | Yes | No | Yes | - | 9.93 |
| 13 | Age estimation | Yes | Yes | No | - | 8.28 |
| 14 | Age estimation | Yes | No | No | - | 8.39 |
| 15 | Age estimation | No | Yes | No | - | 9.68 |
| 16 | Sex estimation | Yes | No | No | 69.64% (70.95%) | - |
| 17 | Sex estimation | Yes | No | Yes | 89.03% (86.73%) | |

Table 2. Comparison of machine learning model accuracy (ResNet101) with different combinations of analysis using the Walker-Data. MAE is the mean age error, that is plus or minus x years. Scenario 16 is conventional landmark analysis based on coordinates subject to Generalised Procrustes Analysis (GPA) so including only elements of shape. The figure in parentheses is the jack-knifed result. Scenario 17 is conventional landmark analysis based on inter-landmark distances so includes elements of shape and size. The figure in parentheses is the jack-knifed result.

**Figures**

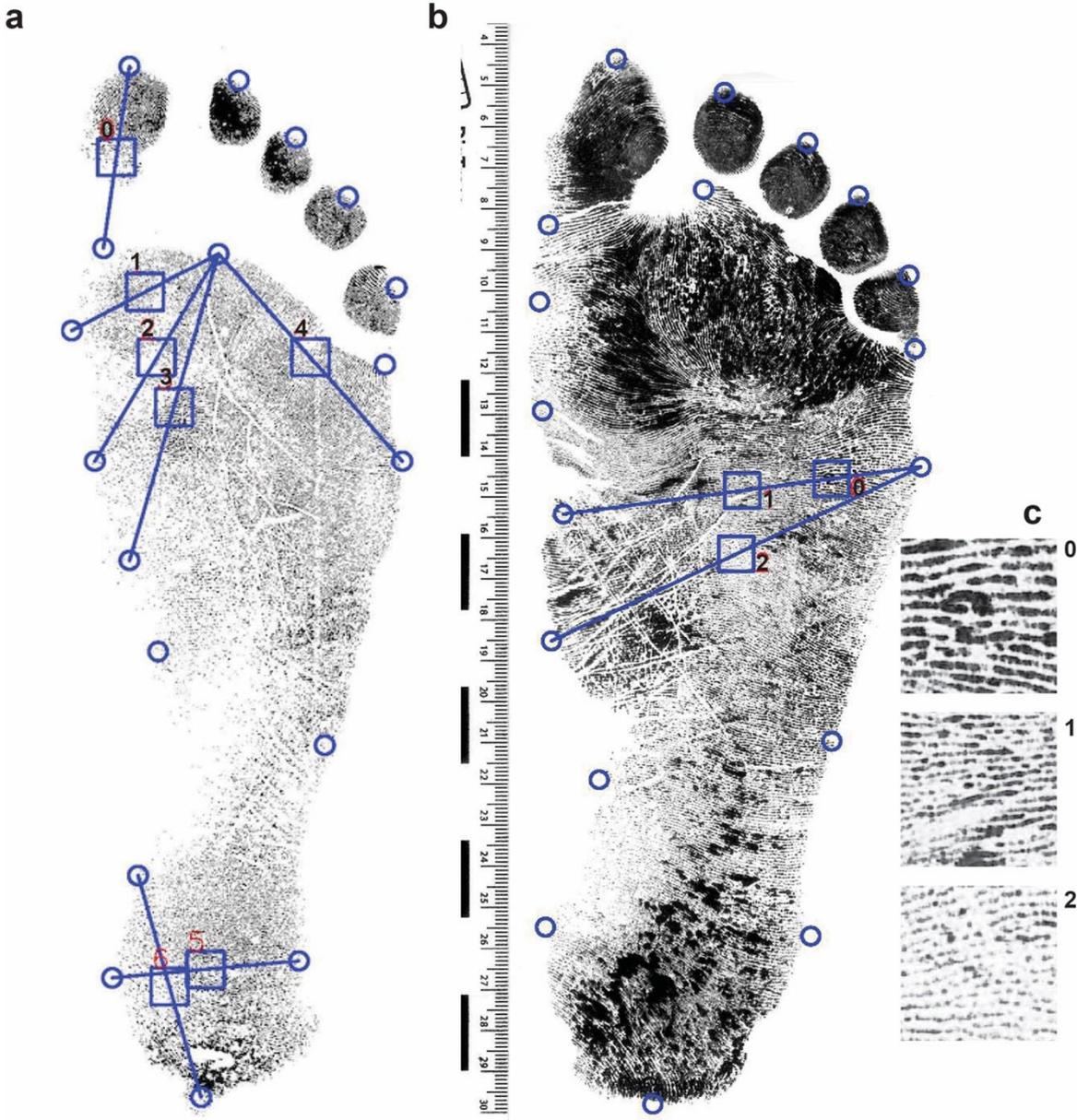

Figure 1: Landmarks and sampling areas. a. Initial sampling points mid-way between the placed landmarks. b. Revised and additional landmarks. c. Cropped sample areas.

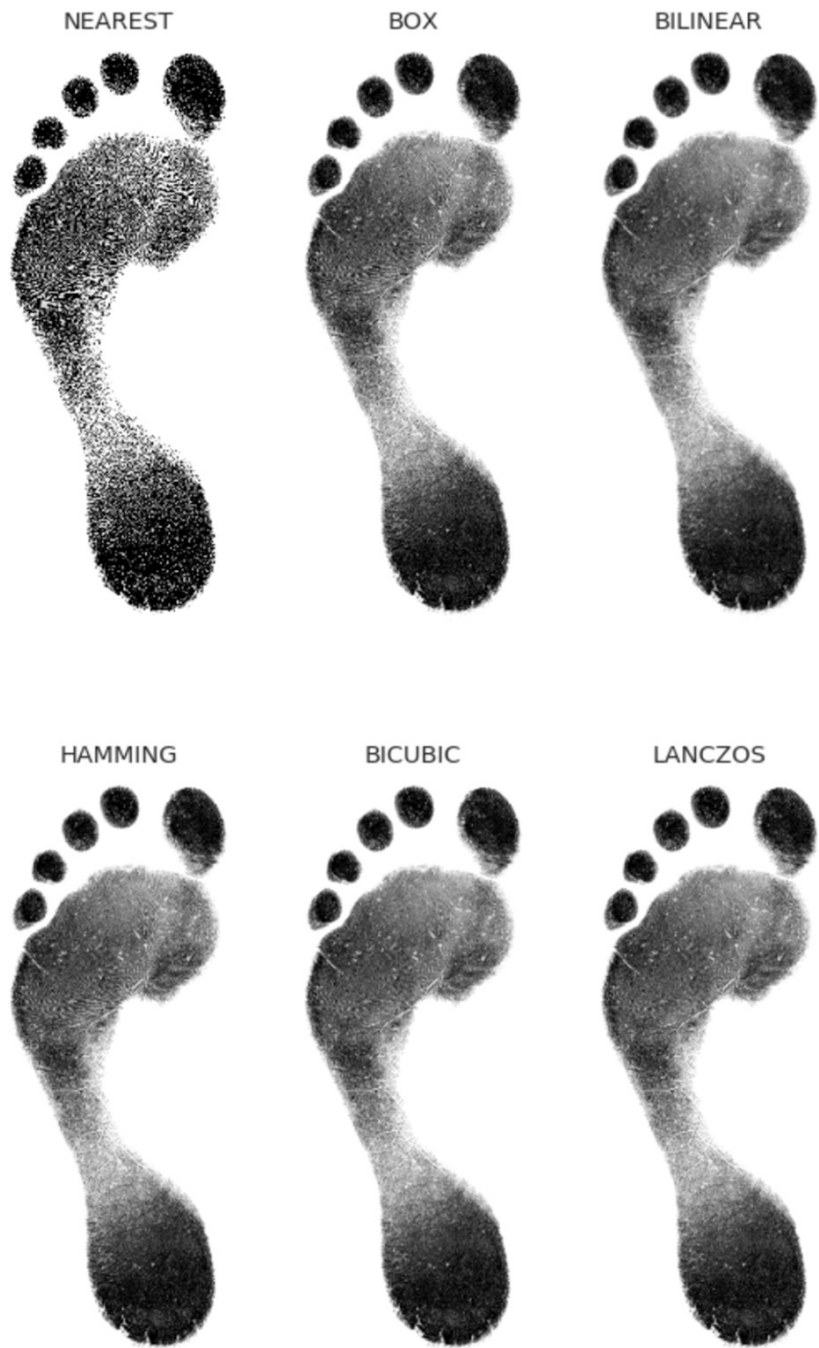

Figure 2: The results of different resizing algorithms.

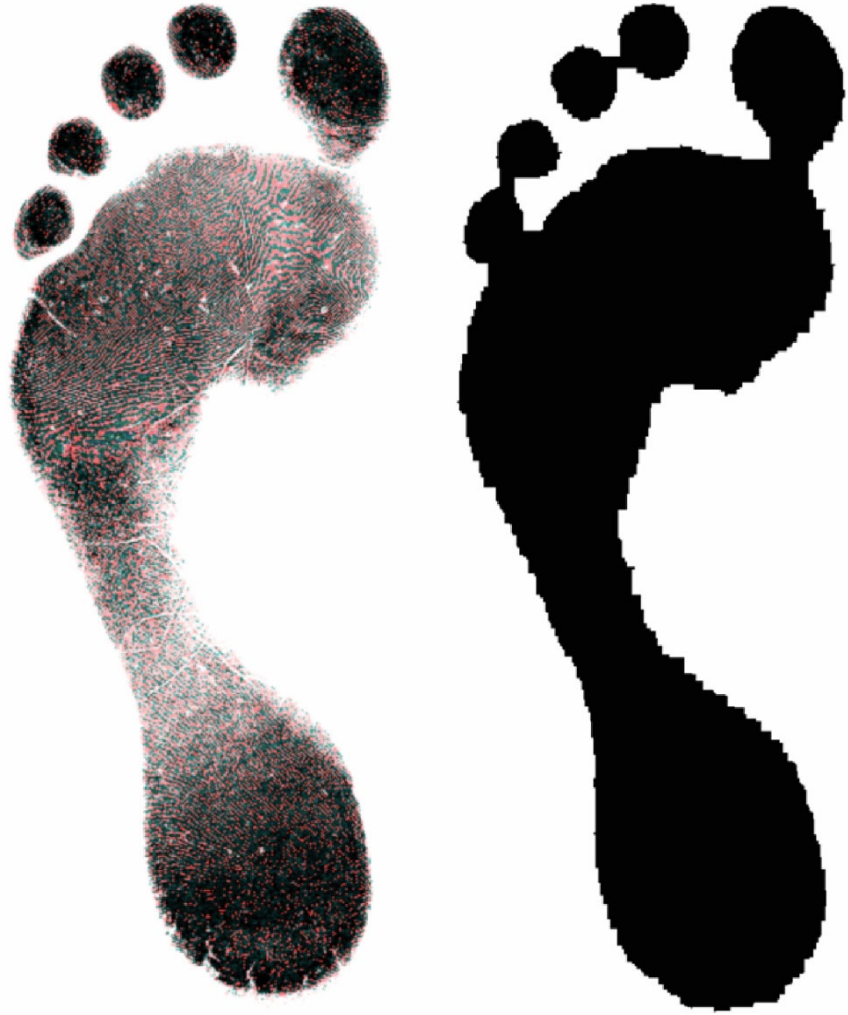

Figure 3: Image analysis. a. Image with channels composed using 3 resizing methods (R=NEAREST, G=BILINEAR, B=HAMMING). B. The same print with ridge detail removed using dilation operation followed by erosion operation.

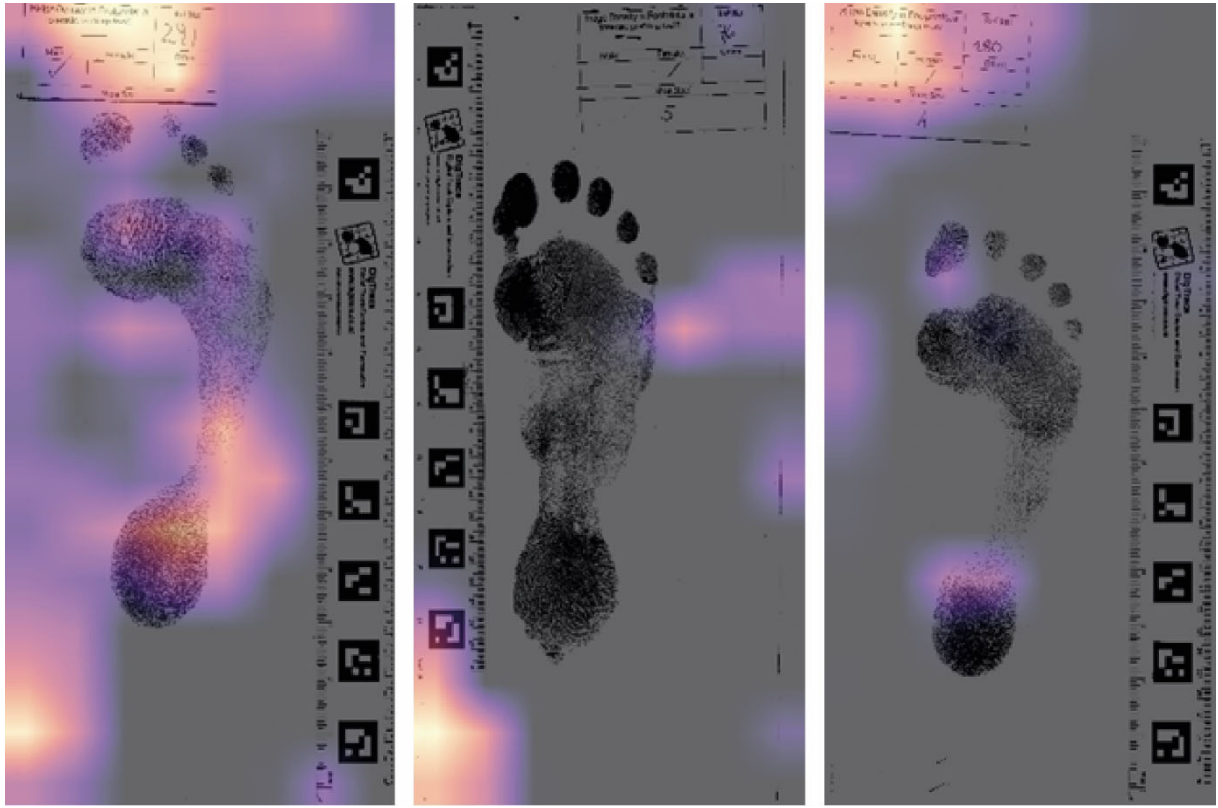

Figure 4: Example of the machine learning output where the focus for decision-making as indicated by the heat map is on the record card which has a sex indicator on it.

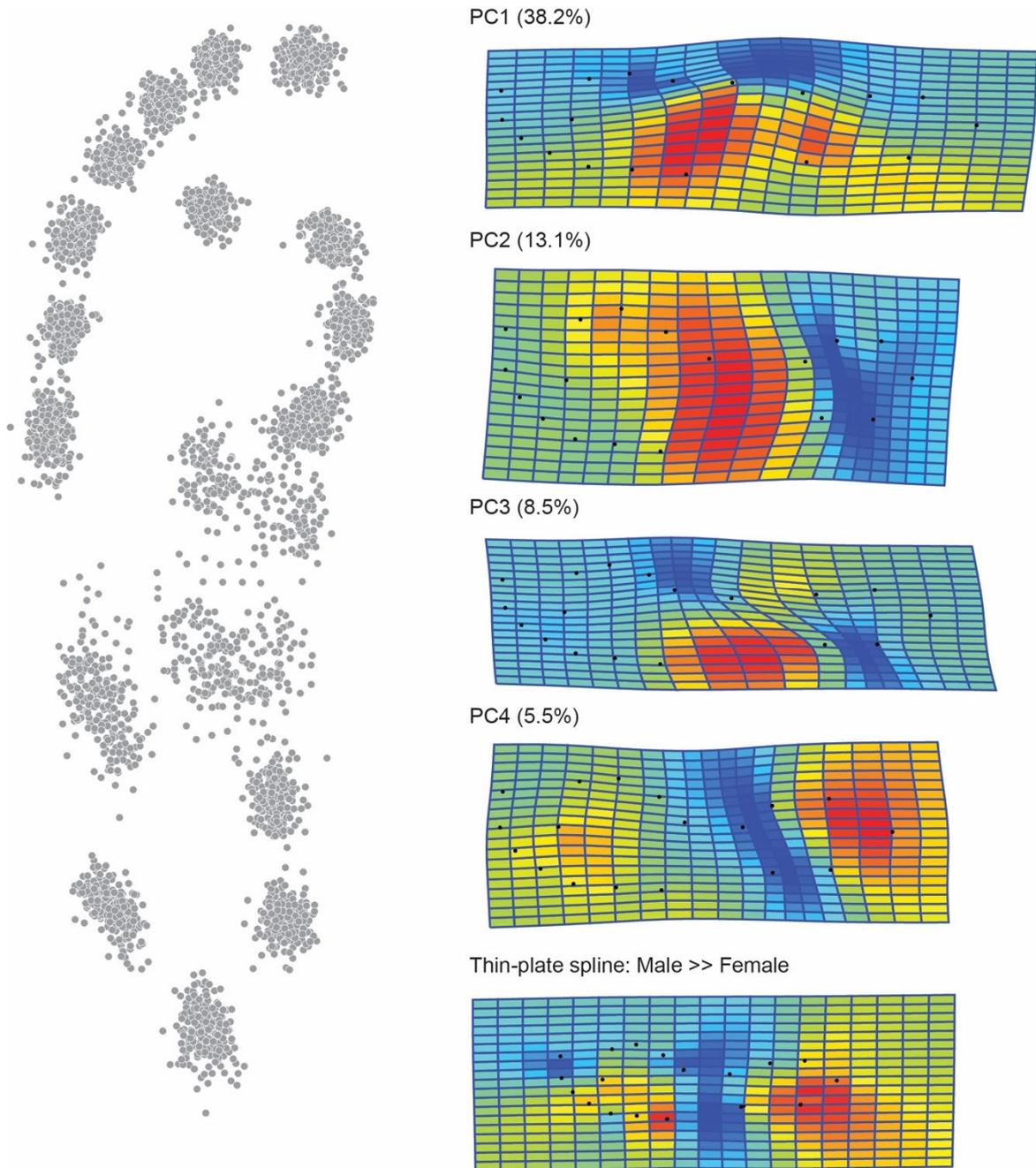

Figure 5: Geometric morphometrics on the Bournemouth-Data, showing the distribution of landmarks, the first four Principal Component Warps (65.3% of the variance) and thin-plate spline comparison of the male to female means. N= 132 males and 165 females.

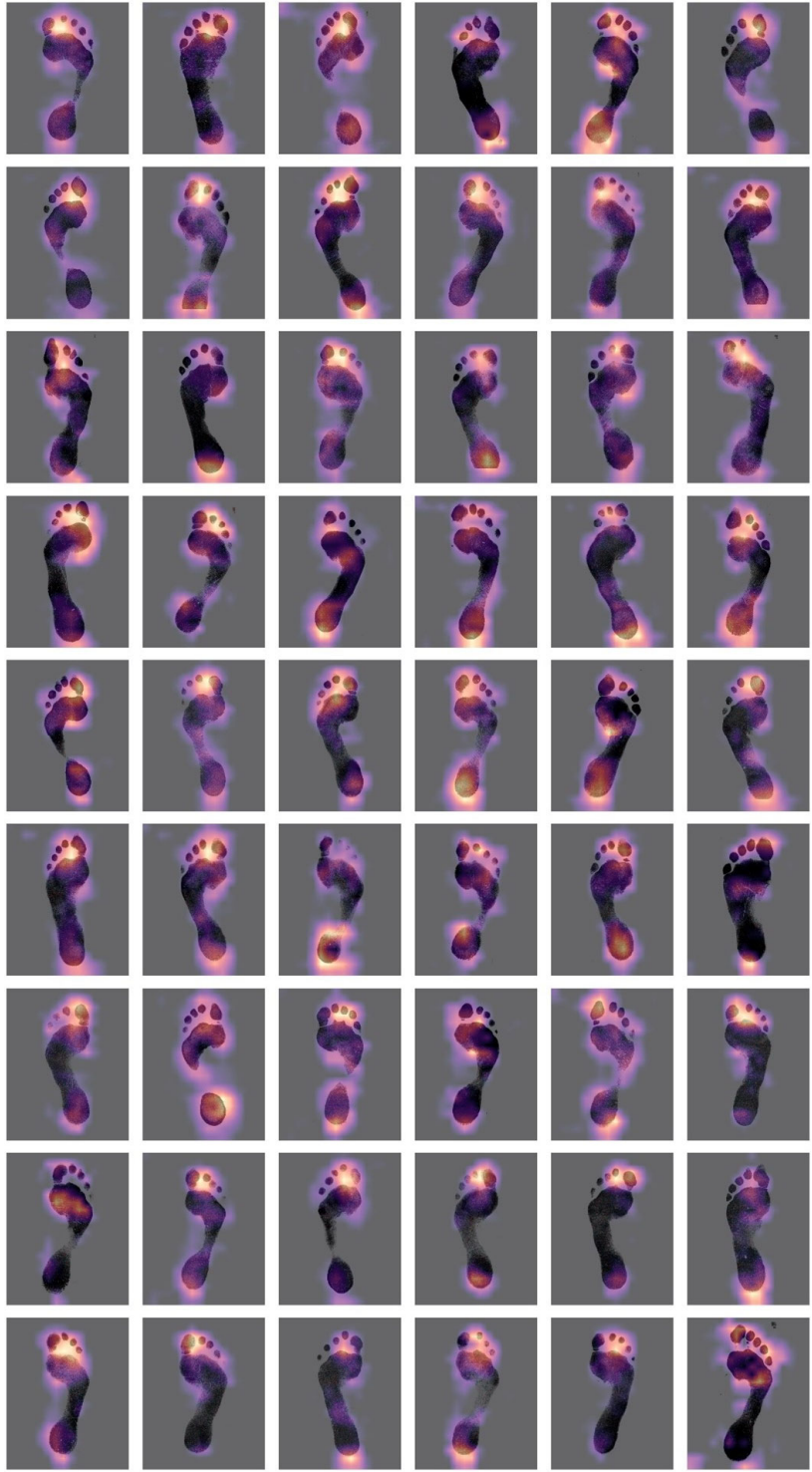

Figure 6: Grad-CAM heatmaps for textured inputs with the lowest entropy losses. The heat maps indicate areas which a key in the decision-making process. Note the concentration of bright areas around the heel and just behind the first and second toes.